\documentclass{article}

\pdfoutput=1
\usepackage{arxiv}

\usepackage[utf8]{inputenc} 
\usepackage[T1]{fontenc}    
\usepackage{hyperref}       
\usepackage{url}            
\usepackage{booktabs}       
\usepackage{amsfonts}       
\usepackage{nicefrac}       
\usepackage{microtype}      
\usepackage{lipsum}		
\usepackage{graphicx}
\usepackage{subfigure}
\usepackage{natbib}
\usepackage{doi}
\usepackage{booktabs}
\usepackage[table,xcdraw]{xcolor}
\usepackage{amsfonts}%
\usepackage{amsmath}%
\usepackage{amssymb}%
\usepackage{graphicx}
\usepackage{algorithm}
\usepackage{algorithmic}

\providecommand{\U}[1]{\protect\rule{.1in}{.1in}}
\newtheorem{theorem}{Theorem}

\newtheorem{proposition}[theorem]{Proposition}

\title{AsymptoticNG: A Regularized Natural Gradient Optimization Algorithm with Look-ahead Strategy}

\author{ 
	{\hspace{1mm}Zedong~Tang,~Fenlong~Jiang,~Junke Song,~Maoguo~Gong,~Hao~Li} \\
	School of Electronic Engineering\\
	Xidian University\\
	Xi'an, China \\
	\texttt{gong@ieee.org}\\
	\And
	{\hspace{1mm}Fan~Yu,~Zidong~Wang,~Min~Wang} \\
	Central Software Institute of 2012 Lab\\
	Huawei Technologies Co. Ltd\\
	Hangzhou, China \\
	\texttt{fan.yu@huawei.com} \\
}

\renewcommand{\undertitle}{}
\renewcommand{\shorttitle}{AsymptoticNG}

\hypersetup{
pdftitle={A template for the arxiv style},
pdfsubject={q-bio.NC, q-bio.QM},
pdfauthor={David S.~Hippocampus, Elias D.~Striatum},
pdfkeywords={First keyword, Second keyword, More},
}

\begin{document}
\maketitle

\begin{abstract}
	Optimizers that further adjust the scale of gradient, such as Adam, Natural Gradient (NG), etc., despite widely concerned and used by the community, are often found poor generalization performance, compared with Stochastic Gradient Descent (SGD). They tend to converge excellently at the beginning of training but are weak at the end. An immediate idea is to complement the strengths of these algorithms with SGD. However, a truncated replacement of optimizer often leads to a crash of the update pattern, and new algorithms often require many iterations to stabilize their search direction. Driven by this idea and to address this problem, we design and present a regularized natural gradient optimization algorithm with look-ahead strategy, named asymptotic natural gradient (ANG). According to the total iteration step, ANG dynamic assembles NG and Euclidean gradient, and updates parameters along the new direction using the intensity of NG. Validation experiments on CIFAR10 and CIFAR100 data sets show that ANG can update smoothly and stably at the second-order speed, and achieve better generalization performance.
\end{abstract}


\section{Introduction}
	The optimization algorithms represented by Stochastic Gradient Descent (SGD) support and promote the rapid development of deep learning in many fields, including but not limited to computer vision\citep{he2016deep, ren2015faster}, natural language processing\citep{vaswani2017attention}, and machine reasoning\citep{silver2016mastering}. Evaluated on a minibatch, SGD updates parameters in the model along the negative gradient direction with uniform scale. While generally simple and effective, it is often necessary to laboriously tune the hyper-parameters for some large tasks, for example, to reduce the learning rate in stages during the training process\citep{goyal2017accurate}. 
	
	To tackle this problem, a number of adaptive variants have emerged, such as Adagrad\citep{duchi2011adaptive}, Adadelta\citep{zeiler2012adadelta}, RMSprop\cite{tieleman2012lecture}, Adam\citep{kingma2014adam}. These methods often diagonally scale the gradient by estimating the curvature of loss function. This kind of precondition for the gradient of each parameter, in a sense, realizes the self-adaptive learning rate adjustment. However, their effectiveness can only be assured if the model is well-conditioned, that is, the principal directions of curvature should better be aligned with the coordinates of the parameters \citep{arbel2019kernelized}.
	
	Taking into account the correlation of different parameters, the second-order optimization algorithms can often give a more accurate estimate of curvature. In general, given a loss function $\ell$, the second-order optimizers update the parameters $\theta$ as: $\theta  \leftarrow \theta  - \eta {G^{ - 1}}\nabla \ell $, where $\eta$ is a positive learning rate and $G^{ - 1}$ is a preconditioner to scale the gradient $\nabla \ell $, which captures the local curvature or related information on Euclidean space or the manifold of probability distribution. For the latter, we generally call it Fisher information matrix, and then ${G^{ - 1}}\nabla \ell $ defines the natural gradient (NG)\citep{amari1998natural}, which represents the steepest direction of objective function in the parameter space, as measured by the KL-divergence. Although NG's effectiveness has been verified on some toy models, the calculation will become infeasible as the number of model parameters (${n_\theta }$) increases, because matrix G has a gigantic size ${n_\theta } \times {n_\theta }$ and it is too large to compute and invert it directly. Therefore, It is necessary to make a trade-off between the quality of curvature information and computational efficiency. This stimulated a large number of methods for estimating NG. Such algorithms often usually take advantage of the hierarchical structure of neural networks to propose a more convenient form of computation\citep{heskes2000natural, martens2015optimizing, ollivier2015riemannian, povey2014parallel, roux2007topmoumoute}.
	
	Owing to competitive performance and lightweight parameter tuning, adaptive and second-order algorithms have been utilized in many applications. Recently, however, it has been widely reported in the community that the final generalization of these algorithms is often not as good as SGD\citep{wilson2017marginal, reddi2019convergence, keskar2017improving, anil2020second, osawa2019large}. For adaptive optimizers, several approaches have been proposed to mitigate this phenomenon. \citet{zhang2018improved} proposed normalized direction-preserving Adam, which can preserve the gradient direction for updating weight vectors. Similarly, but not introducing additional hyperparameters, SWATS introduces a method for smoothly transitioning from Adam to SGD\citep{keskar2017improving}. 
	
	Unfortunately, there are few methods to analyze this similar phenomenon in second-order algorithms and thus make improvements. In order to fill the gaps in the research, in this paper, we choose Kronecker-factored Approximate Curvature (K-FAC) \citep{martens2015optimizing} as the NG estimation method, and propose a regularized natural gradient optimization algorithm with look-ahead strategy, named asymptotic natural gradient (ANG), which dynamically mixes the normalized Natural gradient direction with the Euclidean gradient direction. The whole process is very smooth and there is no mode oscillation caused by truncated transformation. In addition, aiming at the problem that inverse is difficult in second-order algorithms, we further proposed a inverse-free ANG (IFANG), which is more device-friendly. It further approximates the factor matrix in K-FAC, and utilizes SMW formula to reduce the inverse operation to the inverse of a scalar. This is a lighter and faster alternative.

\section{Look Ahead Strategy in Natural Gradient Optimization}
In this section we propose an estimator with look-ahead strategy, which adds a
novel regularizer in the equation (1). We first start by presenting the
look-ahead strategy by maximating the model change in the ahead area of the
model space. Then, for reducing the computational complexity of calucating the
natural gradient, the alternative low-rank inverse method is proposed. In
section, we discuss the practical issue in details.

\subsection{Look Ahead Strategy}

We start by giving an intuitive view on the proposed look-ahead strategy. We
say that in the neighbor region around the current model, the search space is
flat, namely, the gradients of parameters are similar. Therefore, the
look-ahead strategy allows the optimizer moves a step along the direction $u$
that needs to calculate. Then, the optimizer is expect to move a step in the
predicted position along the direction of the negative gradient, because
without any prior the loss changes steeply along this direction. And the model
reaches a test position. We measure the corresponding model change in the
parameterized space at the test point. Unlike the regularizer $u^{T}Gu$, which limits the model
change in the parameterized space to keep the proper approximation, we expect
to maximum the model change in the futher test position, such that the
optimizer can get rid of the flat aera and saddle-point by moving forward the
variable region.

We formulate the strategy by the following new estimator.

\begin{proposition}
	Asumme that $\lambda_{1}>0$ and $\lambda_{2}>0$, then the look-ahead
	regularizer is give by:%
	\begin{equation}
		\begin{aligned}
			\nabla^{D}L(\theta)  =\arg\min_{u\in%
				\mathbb{R}
				^{q}}M(u)+\frac{\lambda_{1}}{2}u^{T}Gu
			 -\frac{\lambda_{2}}{2}\left(  u-\nabla L(\theta)\right)  ^{T}G\left(
			u-\nabla L(\theta)\right)  ,
		\end{aligned}
		\label{eq_estimator}
	\end{equation}
	where $\lambda_{1}$ and $\lambda_{2}$ are two parameters to balance the
	influence of two regularizers.
\end{proposition}

In equation (\ref{alg_ang}), $M(u)$ is the linear approximation of $L(x)$ near $x$. The first regularizer $\frac{\lambda_{1}}{2}u^{T}Gu$ is measured the model
change by $G$, which keeps a good approximation in the neighbor region and constraints the model change with respect to the parameter update. The second regularizer $-\frac{\lambda_{2}}{2}\left(  u-\nabla
L(\theta)\right)  ^{T}G\left(  u-\nabla L(\theta)\right)$ is to make sure
that the futher model change is maximized after moving toward the steep descent
direction indicated by the negative gradient $-\nabla L(\theta)$. In the other word, the combined search direction $u-\nabla L(\theta)$ generates a large model change. By this means, the optimizer can consider the future search behavior that preserves the fast convergence and escapes from the saddle points. It is note that the negative coefficient makes this term maximum although we minimize the sum of these three terms.

\begin{proposition}
	The natural gradient estimated by equation (\ref{eq_estimator}) is given by:%
	\begin{equation}
		\nabla^{D}L(\theta)=\left(  \frac{1}{\lambda_{1}-\lambda_{2}}G^{-1}%
		+\frac{\lambda_{2}}{\lambda_{1}-\lambda_{2}}I\right)  t,
		\label{eq_ang2}
	\end{equation}
	where $t=-\nabla L(\theta)$.
\end{proposition}

The proposed method takes $\left(  \frac{1}{\lambda_{1}-\lambda_{2}}%
G^{-1}+\frac{\lambda_{2}}{\lambda_{1}-\lambda_{2}}I\right)  $ as the
preconditioner. Equation (3) implies that the look ahead strategy actually
combines the natural gradient and the Eulicude gradient linearly, thus, the
key issue is how to approproately set the additional parameters $\lambda_{1} $
and $\lambda_{2}$. An adaptive method of setting these two parameters is
devised, which is presented in the following section.

\subsection{Spherical Linear Blending}

The equation (\ref{eq_ang2}) can be rewritten as%

\begin{equation}
	\nabla^{D}L(\theta)=\frac{1}{\lambda_{1}-\lambda_{2}}\left(  G^{-1}%
	t+\lambda_{2}t\right)  .
\end{equation}

In order to balance the scale of two terms in above equation, namely the
Fisher natural gradient and Euclidean gradient, we employ following settings
for $\lambda_{1}$and $\lambda_{2}$.%
\begin{equation}
{\lambda _1}{\rm{ = }}\frac{{{{\left\| t \right\|}_G}}}{{\left\| t \right\|}} + 1, 
{\lambda _2}{\rm{ = }}\frac{{{{\left\| t \right\|}_G}}}{{\left\| t \right\|}}.
\end{equation}

Then we have following regularized natural gradient,%

\begin{equation}
	\nabla^{D}L(\theta)=\left\Vert t\right\Vert _{G}\left(  \frac{G^{-1}%
		t}{\left\Vert t\right\Vert _{G}}+\frac{t}{\left\Vert t\right\Vert }\right)  .
\end{equation}

Note that the first term and second term is the normalized natural gradient
and Euclidean gradient. We can view the linear combination of this two vectors
as the the derived search direction, while $\left\Vert t\right\Vert _{G}$ is
the amplitude of regularized natural gradient. By introducting an additional
positive parameter $\lambda\geq0$, we can control the ingredient of two search
directions, which achieves the trade-off between fast convergence and generalization.

\begin{proposition}
	The adaptive version of equation (4) is given by,
\end{proposition}%

\begin{equation}
	\nabla^{D}L(\theta)=\left\Vert t\right\Vert _{G}\left(  (1-\lambda
	)\frac{G^{-1}t}{\left\Vert t\right\Vert _{G}}+\lambda\frac{t}{\left\Vert
		t\right\Vert }\right)  .
\end{equation}

This variant of equation (4) is the linear blending of the two search
directions. The serach direction caluclated by equation (5) is located on the
convex hull of two vectors $\frac{G^{-1}t}{\left\Vert t\right\Vert _{G}}$ and
$\frac{t}{\left\Vert t\right\Vert }$. We expect that the learning process can
draw the balance of convergence and generalization. We say that in the early
stage the natural gradient takes the main role, which benefits the
second-order convergence rate. In the latter stage, the normal gradient draws
a main contribution in the optimization which enhances the generalization of
the trained model. However, the linear combination of $\frac{G^{-1}%
	t}{\left\Vert t\right\Vert _{G}}$ and $\frac{t}{\left\Vert t\right\Vert }$
usually generates a vector with norm smaller than 1. This might underestimate
the search step, which harms the convergence. We employ a smoother strategy
than linear combination, namely spherical linear combination. 

\begin{proposition}
	The spherically adaptive version of equation (4) is given by,
\end{proposition}%

\begin{equation}
	\nabla^{D}L(\theta)   =\left\Vert t\right\Vert _{G}\left(  \frac
	{\sin(1-\lambda)\Omega}{\sin\Omega}\frac{G^{-1}t}{\left\Vert t\right\Vert
		_{G}}+\frac{\sin\lambda\Omega}{\sin\Omega}\frac{t}{\left\Vert t\right\Vert
	}\right), 
	\Omega  =\arccos\frac{t^{T}G^{-1}t}{\left\Vert t\right\Vert _{G}\cdot\left\Vert
		t\right\Vert }.
\end{equation}
It is note that $t^{T}G^{-1}t$ is alway greater than 0. If $t^{T}G^{-1}t$ is
less than 0, the negative gradient is used. This strategy make sure that the
norm of the combination of two vectors equals 1. The parameter $\lambda$ is
set by a linear or exponential descent strategy, which is discussing in the experimental section. Since it is obvious that the proposed method attempts to combine the search direction of natural and Euclidean gradient by a smooth way, we call this method asymptotic natural gradient.

\subsection{Inverse-Free Asymptotic Natural Gradient}

The main bottle-neck of the natural gradient in the practical application is
the calculation of the preconditioner. Thus, in this section, we devise a
computationally efficient method for the proposed inverse-free asymptotic natural gradient method.

The proposed method is based on K-FAC, which can be formulated as follows,
\begin{equation}\widehat{\nabla^{D}L(\theta)}=\left(  \lambda \mathbf{I}+\mathbf{A}\right)  ^{-1}\times-\nabla
	L(\theta)\times\left(  \lambda \mathbf{I}+\mathcal{D}\mathbf{S}\right)  ^{-1}
	\label{eq_K-FAC}.
\end{equation}
Based on the block-diagonal approximation of Fisher matrix in K-FAC, we first derive the low-rank inverse of Fisher matrix. Recall that the independent assumption between activations and output derivatives yields the Kronecker approximation of the Fisher matrix, $\mathbf{G}_\ell=\mathbf{\widehat{\Omega}}_\ell^{(\lambda)} \otimes \mathbf{\widehat{\Gamma}}^{(\lambda)}_\ell$, where $\mathbf{\widehat{\Omega}}^{(\lambda)}_\ell=\lambda \mathbf{I} + \mathbf{\widehat{\Omega}}_\ell$ and $\mathbf{\widehat{\Gamma}}^{(\lambda)}_\ell=\lambda \mathbf{I} + \mathbf{\widehat{\Gamma}}_\ell$, herein, $\lambda$ is the damping introduced by \cite{martens2015optimizing}. 
The unknown data distribution causes the difficulty in computing the Fisher matrix information. To address this difficulty, we estimate the Kronecker factors by the empirical statistics over a mini-batch of training data $\mathcal{B}\subset \mathcal{D}_{training}, |\mathcal{B}|=M$ at iteration $t$.
The empirical statistics of the Kronecker factors over a given mini-batch is defined as
\begin{equation}
	\lambda \mathbf{I} + \mathbf{\widehat{\Omega}}_{\ell} = \lambda \mathbf{I} + \mathbb{E}_\mathcal{B}[ \mathbf{A}_{\ell}^ \top \mathbf{A}_{\ell} ] = (\lambda \mathbf{I} + \frac{\mathbf{A}_{\ell}^\top \mathbf{A}_{\ell}}{M}),
	\label{eq_factor1} 
\end{equation}
\begin{equation}
	\lambda \mathbf{I} + \mathbf{\widehat{\Gamma}_\ell} = \lambda \mathbf{I} + \mathbb{E}_\mathcal{B}[ \mathcal{D}\mathbf{S}_\ell^ \top \mathcal{D}\mathbf{S}_\ell ] = ( \lambda \mathbf{I} + \frac{\mathcal{D}\mathbf{S}^\top_\ell \mathcal{D}\mathbf{S}_\ell}{M}).
	\label{eq_factor2}
\end{equation}
In Section II, we mentioned that $\mathbf{A}$ and $\mathcal{D}\mathbf{S}$ are $\mathbb{R}^{M \times {n_\ell}}$ and $\mathbb{R}^{M \times {n_{\ell-1}}}$, respectively. When $M << {n_\ell},n_{\ell-1}$, they have low-rank structure. The approximation of above terms over a mini batch suggests that the factors of FIM is practically low-rank. Note that although we assume the factors of FIM having low-rank structure, the resulting approximation of FIM is not low-rank. Existing methods \citep{martens2015optimizing,grosse2016kronecker} attempt to reduce the computation cost of the fully exact inverse of the Kronecker factors $\widehat{\mathbf{\Gamma}}^{-1}$ and $\widehat{\mathbf{\Omega}}^{-1}$, where the large matrices have to decompose into the low-rank factors first. In contrast, our method utilizes the low-rank nature of activation likelihood covariance matrix $\widehat{\mathbf{\Omega}}_\ell=\mathbf{A}_\ell\mathbf{A}_\ell^\top$ and pre-activation derivative likelihood covariance matrix $\widehat{\mathbf{\Gamma}}_\ell=\mathcal{D}\mathbf{S}_\ell\mathcal{D}\mathbf{S}_\ell^\top$ matrices to yield an low-rank formulation of inverses of $(\widehat{\mathbf{\Gamma}}^{\lambda}_\ell)^{-1}$ and $(\widehat{\mathbf{\Omega}}^{\lambda}_\ell)^{-1}$.
To reduce the computational, it is reasonable to reduce the first dimension of $\mathbf{A}_\ell$ and $\mathcal{D}\mathbf{S}_\ell$, such that $\mathbf{A}_\ell$ and $\mathcal{D}\mathbf{S}_\ell$ reduce to $s'\times n_\ell$ and $s'\times n_{\ell-1}$, $s'<s$. We first group $M$ rows of $\mathbf{A}_\ell$ and $\mathcal{D}\mathbf{S}_\ell$ into $s'$ group, then the mean of each group is calculated. Thus, $s'$ rows of $n_\ell$-dimension vectors are given. In inverse-free we set $s'=1$ to avoid the matrix inverse.
\begin{proposition}
	Two inverse-free Kronecker factors of the inverse of Fisher information matrix is derived as
	\begin{equation}
		[\widehat{\mathbf{\Omega}}^{(\lambda)}_\ell]^{-1} = (\lambda \mathbf{I} + \widehat{\mathbf{\Omega}}_\ell)^{ - 1} = \frac{1}{\lambda} \mathbf{I} - \frac{1}{\lambda} \mathbf{A}_\ell^\top \omega^*_\ell \mathbf{A}_\ell,
		\label{eq_smw1}
	\end{equation}
	\begin{equation}
		[\widehat{\mathbf{\Gamma}}^{(\lambda)}_\ell]^{-1} = (\lambda \mathbf{I} + \widehat{\mathbf{\Gamma}}_\ell)^{ - 1} = \frac{1}{\lambda} \mathbf{I} - \frac{1}{\lambda} \mathcal{D}\mathbf{\tilde{S}}_\ell^\top \gamma^*_\ell \mathcal{D}\mathbf{\tilde{S}}_\ell,
		\label{eq_smw2}
	\end{equation}
	where $\omega^*_\ell = {{(  \lambda  + \mathbf{\tilde{A}}_\ell \mathbf{\tilde{A}}_\ell^ \top )}^{ - 1}}$, and $\gamma^*_\ell={{(  \lambda + \mathcal{D}\mathbf{\tilde{S}}_\ell \mathcal{D}\mathbf{\tilde{S}}_\ell^ \top )}^{ - 1}}$.
	\label{pro_inversefree}
\end{proposition}

Using Proposition \ref{pro_inversefree}, we get following equation,
\begin{equation}
	\mathbf{G}^{-1}_\ell = \lambda ^{ - 1}[\mathbf{I} - \mathbf{\tilde{A}}_{\ell-1}^ \top (\lambda + \mathbf{ \tilde{A}}_{\ell-1} \mathbf{ A}_{\ell-1}^ \top )^{ - 1} \mathbf{\tilde{A}}_{\ell-1}]
	\otimes \lambda ^{ - 1} [ \mathbf{I} - \mathcal{D}\mathbf{\tilde{S}}_\ell^ \top (\lambda + \mathcal{D}\mathbf{\tilde{S}}_\ell \mathcal{D}\mathbf{ \tilde{S}}_\ell^ \top )^{ - 1} \mathcal{D}\mathbf{\tilde{S}}_\ell].
\end{equation}

\begin{proposition}
	The layer-wise inverse-free asymptotic natural gradient is given as,
	\begin{equation}\widehat{\nabla^{D}L(\theta_\ell)}=[\widehat{\mathbf{\Omega}}^{(\lambda)}_\ell]^{-1}\times-\nabla
		L(\theta_\ell))\times[\widehat{\mathbf{\Gamma}}^{(\lambda)}_\ell]^{-1}
		\label{eq_if}.
	\end{equation}
\end{proposition}

Taking $[\widehat{\mathbf{\Gamma}}^{(\lambda)}_\ell]^{-1}$ as an example, the computation cost of inverting the factors of FIM reduces from $O(n_\ell^3)$ to roughly $O(1)$, where $n_\ell$ is the number of outputs of the $l$th layer. Our derived inverse-free procedure is much more efficient than inverting the full Kronecker factors directly. Compared with K-FAC, the theoretical computational cost of computing the inverse of Kronecker factors of FIM is reduced from $O(\sum_{\ell}n_\ell^3 + \sum_\ell{n_\ell^2})$ to $O(L + \sum_\ell{n_\ell^2})$. By parallel computing, the matrix multiplication in equations (\ref{eq_smw1})(\ref{eq_smw2}) has a sublinear complexity $O(\sum_\ell{({n_\ell^2/N_T})^2})$, where $N_T$ is the number of computation threads. The main steps is shown in Algorithm \ref{alg_ang}.

\begin{algorithm}[htb]
	\caption{Calculate the terms $[\widehat{\mathbf{\Gamma}}^{(\lambda)}_\ell]^{-1}$ or $[\widehat{\mathbf{\Omega}}^{(\lambda)}_\ell]^{-1}$.}
	\label{alg_factor}
	\begin{algorithmic}[1]
		\REQUIRE layer $\ell$, batch size $M$, $\ell$'s input $\mathbf{A}_{\ell} \in \mathbb{R}^{M \times n_{\ell}}$, $\ell$'s pre-activation derivations $\mathcal{D}\mathbf{S}_\ell \in \mathbb{R}^{M \times n_\ell}$, damping coefficient $\lambda$
		\ENSURE the inverse of damped factor matrices: $[\widehat{\mathbf{\Omega}}^{(\lambda)}_\ell]^{-1}$ and $[\widehat{\mathbf{\Gamma}}^{(\lambda)}_\ell]^{-1}$
		\FOR{$\mathbf{Y}$, $\mathbf{X}$ in $[\{\mathbf{A}_{\ell},~\widehat{\mathbf{\Omega}}^{(\lambda)}_{\ell}\},~\{\mathcal{D}\mathbf{S}_\ell,~\widehat{\mathbf{\Gamma}}^{(\lambda)}_\ell\}]$}
		\STATE $M,n = shape(\mathbf{Y})$.
		\STATE Apply mean reduce in the first dimension of $\mathbf{Y}$
		\IF{$M \ge n$}
		\STATE $\mathbf{X}^{ - 1} = (\lambda \mathbf{I} + \mathbf{Y}^ \top \mathbf{Y})^{ - 1}$
		\ELSE
		\STATE $\mathbf{X}^{ - 1} = \lambda ^{ - 1}[\mathbf{I} - \mathbf{Y}^ \top (\lambda + \mathbf{Y}\mathbf{Y}^ \top )^{ - 1}\mathbf{Y}]$
		\ENDIF
		\ENDFOR
	\end{algorithmic}
\end{algorithm}

\begin{algorithm}[htb]
	\caption{ANG}
	\label{alg_ang}
	\begin{algorithmic}[1]
		\STATE \textbf{parameters:} learning rate $\eta$, damping $\lambda$
		\STATE Initialize $\theta_0$.
		\FOR{$t=1,\cdots,T$}
		\FOR{$\ell=1,\cdots,L$}
		\STATE receive gradient $\nabla L(\theta_\ell^{(t-1)})$.
		\STATE receive $\mathbf{A}_\ell$ and $\mathcal{D}\mathbf{S}_\ell$
		\STATE calculate the factors $[\widehat{\mathbf{\Gamma}}^{(\lambda)}_\ell]^{-1}$ and $[\widehat{\mathbf{\Omega}}^{(\lambda)}_\ell]^{-1}$ via \textbf{Algorithm \ref{alg_factor}}.
		\STATE set natural gradient $\widehat{\nabla^DL(\theta_\ell^{(t-1)})}$ by Proposition 6
		\STATE update $\theta_\ell^{(t)}\leftarrow\theta_\ell^{(t-1)}+\eta\widehat{\nabla^DL(\theta_\ell^{(t-1)})}$
		\ENDFOR
		\ENDFOR
	\end{algorithmic}
\end{algorithm}

\section{Experiments}
To demonstrate the effectiveness of our approach, we conducted extensive experiments on CIFAR10 and CIFAR100 data sets using the ResNet model with two representative optimizers: SGD and K-FAC. And all the experiments are performed on a device equipped with a single TITAN RTX using the Pytorch framework. The experiment was designed into two parts. In part one, we first compare the training and validation performances of SGD, K-FAC, truncated K-FAC (switch K-FAC to SGD in one epoch), and the proposed ANG. Then, we analyze the impact of different $\lambda$ parameter strategies on ANG performance. In the second part, we discussed the performance of INANG, especially its advantages in iteration speed. These experiments are described in detail in the following two sections.

\subsection{Part One}
\begin{figure}[htbp]

	\centering

	\subfigure[{Validation Loss vs. Epoch}]{
		\includegraphics[scale=0.5]{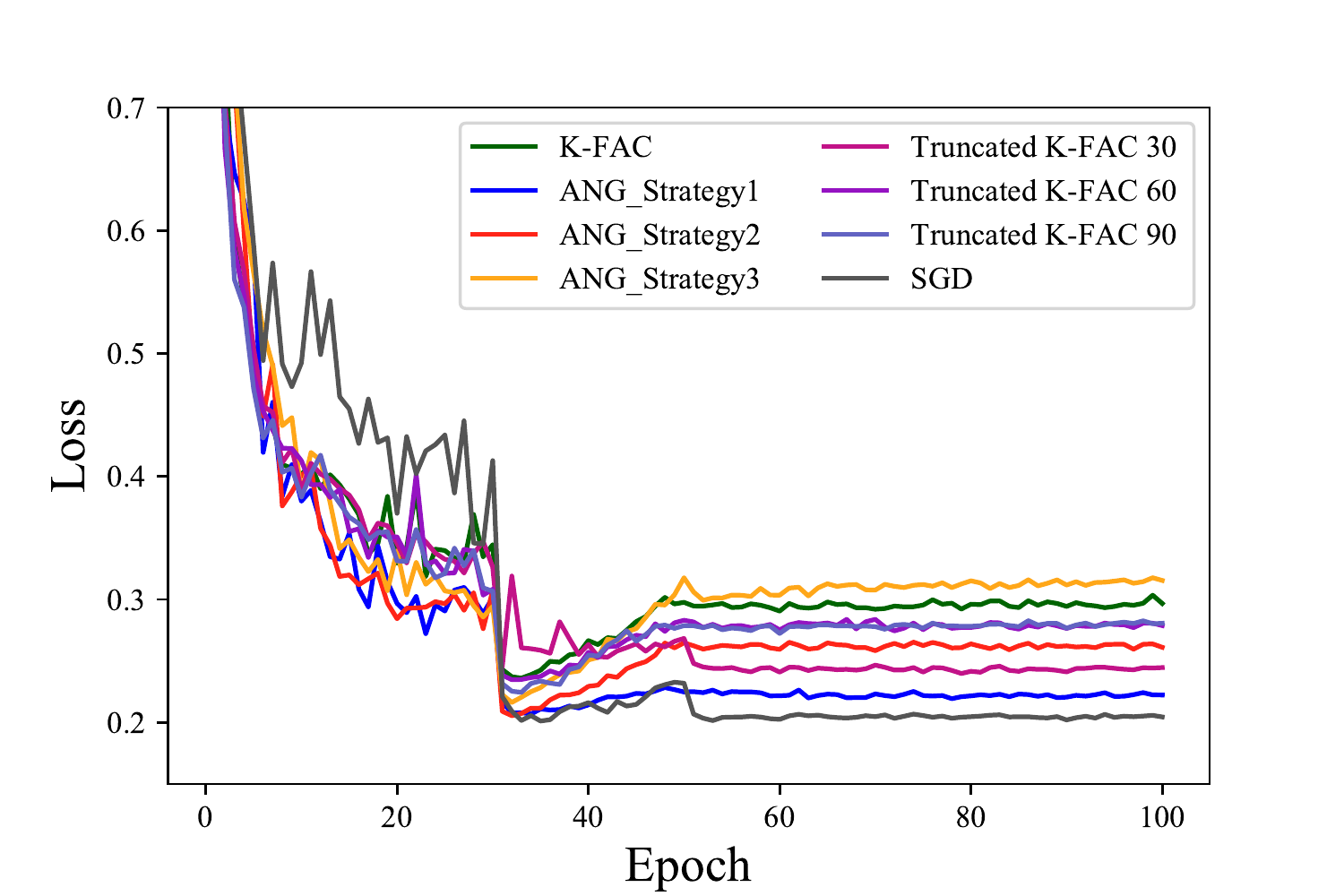}
	}
	\subfigure[{Validation Accuracy vs. Epoch}]{
	\includegraphics[scale=0.5]{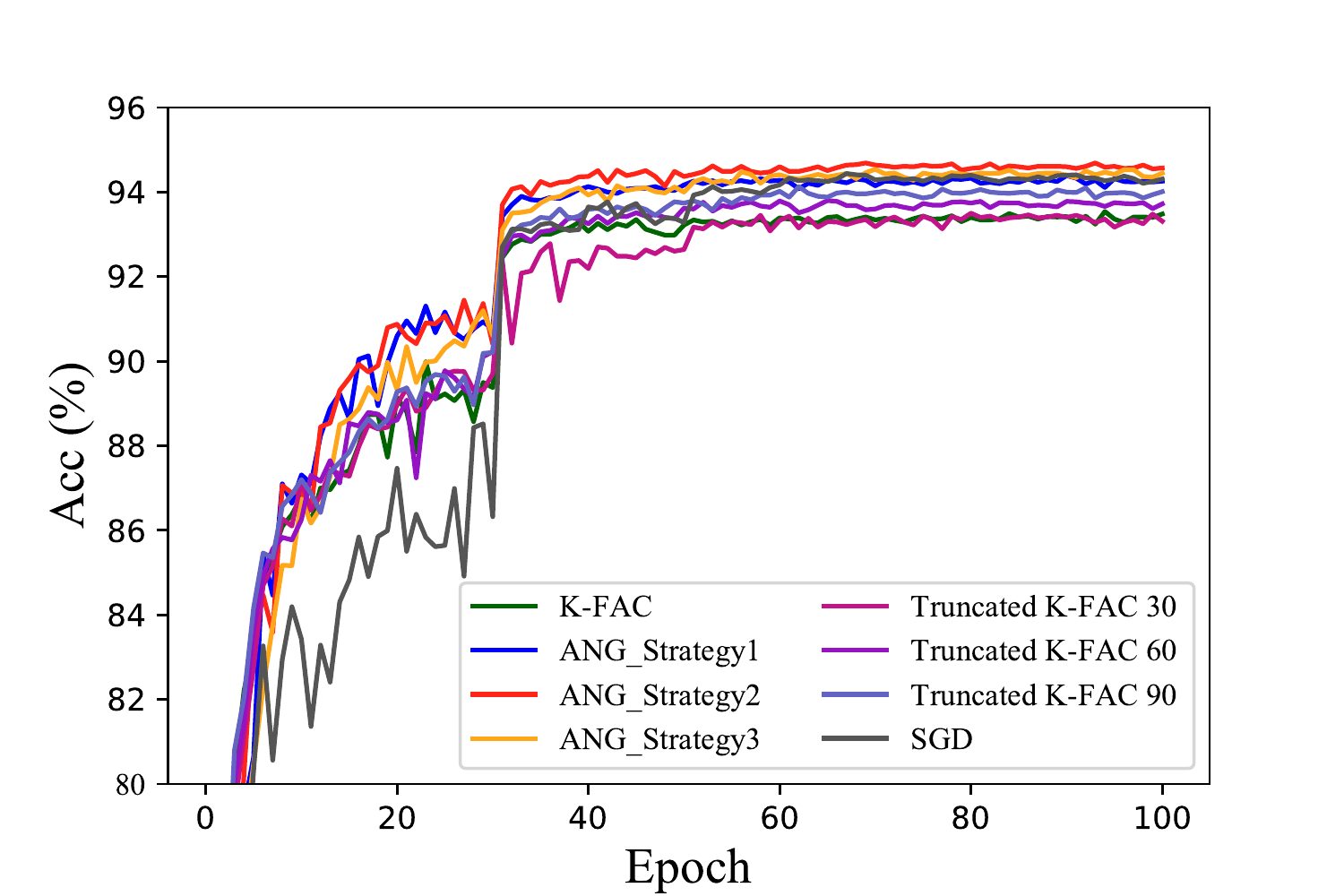}
}
	\subfigure[{Validation Loss vs. Epoch}]{
	\includegraphics[scale=0.5]{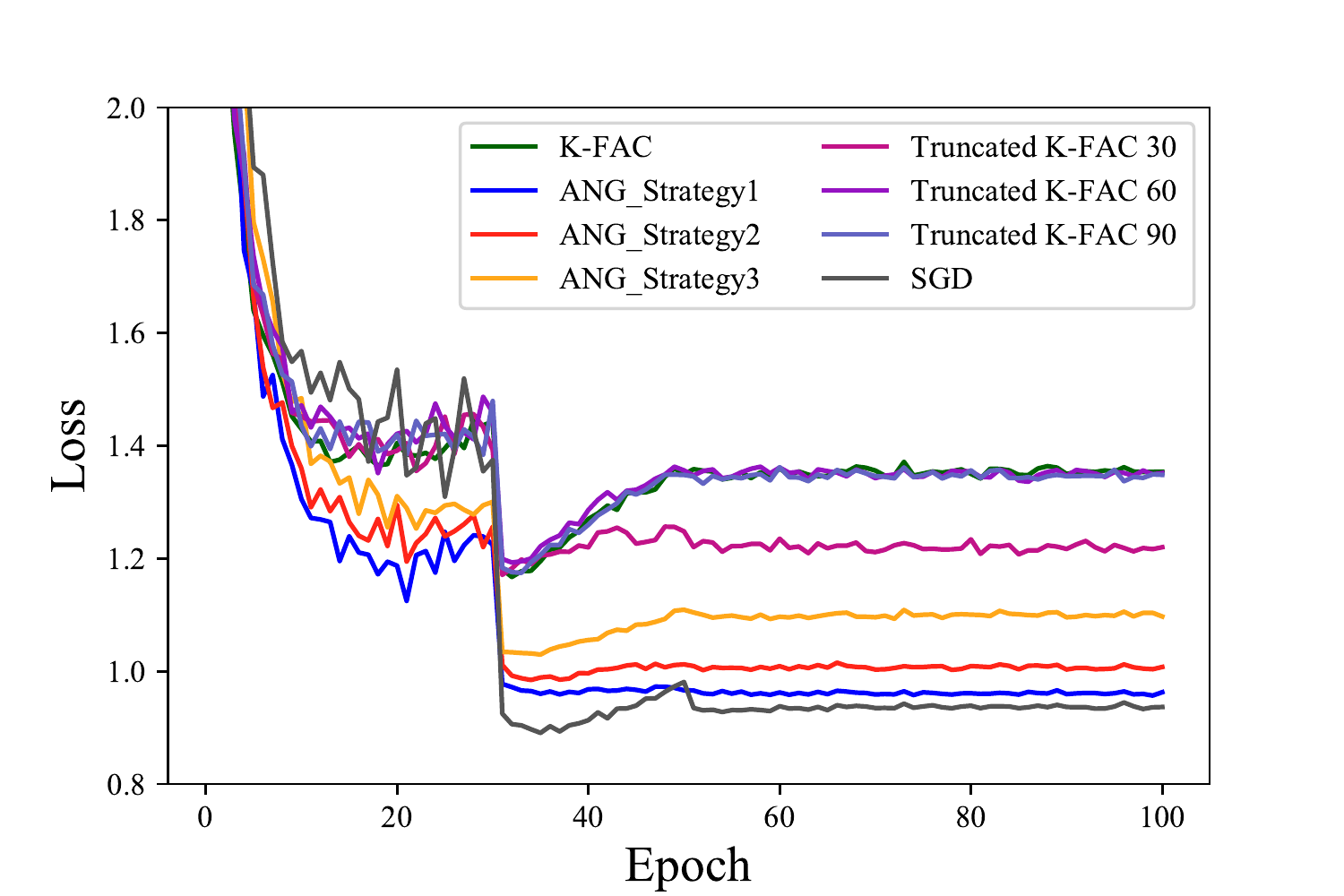}
}
	\subfigure[{Validation Accuracy vs. Epoch}]{
		\includegraphics[scale=0.5]{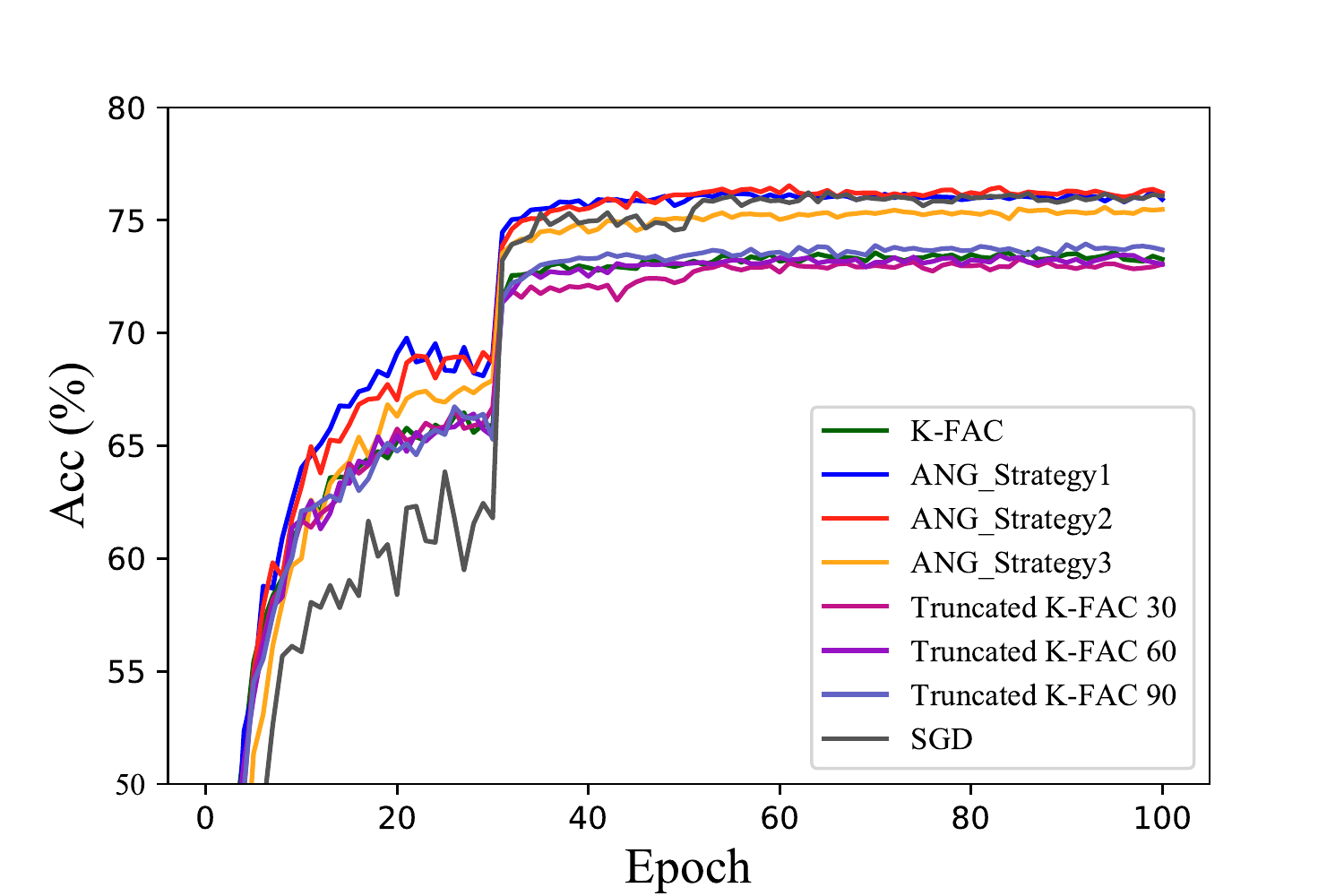}
	}

	\caption{{Performance comparison among SGD, K-FAC, ANG and Truncated K-FAC on CIFAR-10 and CIFAR-100. \textbf{(a)}, \textbf{(b)} show the accuracy and loss curves on CIFAR-10 respectively, and \textbf{(c)}, \textbf{(d)} on CIFAR-100. In the legend, we name ANG with different strategies as ANG\_strategy1, ANG\_strategy2 and ANG\_strategy3. Three curves of switching K-FAC to SGD at 30, 60 and 90 epoch are named as Truncated K-FAC 30, 60 and 90 respectively.} 
	}
\label{fig:fig1}
\end{figure}

In this section, we compare our proposed methods with two baselines of first-order SGD and second-order K-FAC. In addition, we designed several extra experiments to demonstrate the stability and efficiency of our algorithm, as compared to switching K-FAC to SGD truncated in an epoch (30/ 60/ 90). We deploy the learning rate of each method with the best performance on average accuracy of methods in comparisons. Specifically, the initial learning rate of SGD was set to 0.01, while that of other methods was set to 0.005, and they all had a milestone step-down during the training. Results are illustrated in Fig. \ref{fig:fig1}. It can be seen from it that in the two data sets, the final generalization accuracy of K-FAC is not as good as that of SGD under the condition of early rapid convergence. This is also reflected in the Loss curve. SGD effectively inhibits the occurrence of overfitting in the later stage, while K-FAC shows the phenomenon of overfitting, which first decreases and then increases, and the generalization performance is poor. In the three results of truncated K-FAC, oscillation phenomena appear to a certain extent respectively. Moreover, the later the switching is, the lighter the oscillation is, and the higher the final generalization accuracy is. This also indicates that SGD can provide better convergence direction at the end. However, ANG achieves the expected effect. It fisrt converges rapidly at a second-order speed, while in the later stage, it converges close to the first-order generalization performance. The transformation is very smooth without any oscillation.

\begin{figure}[htbp]
	\centering
	\includegraphics[scale = 0.6]{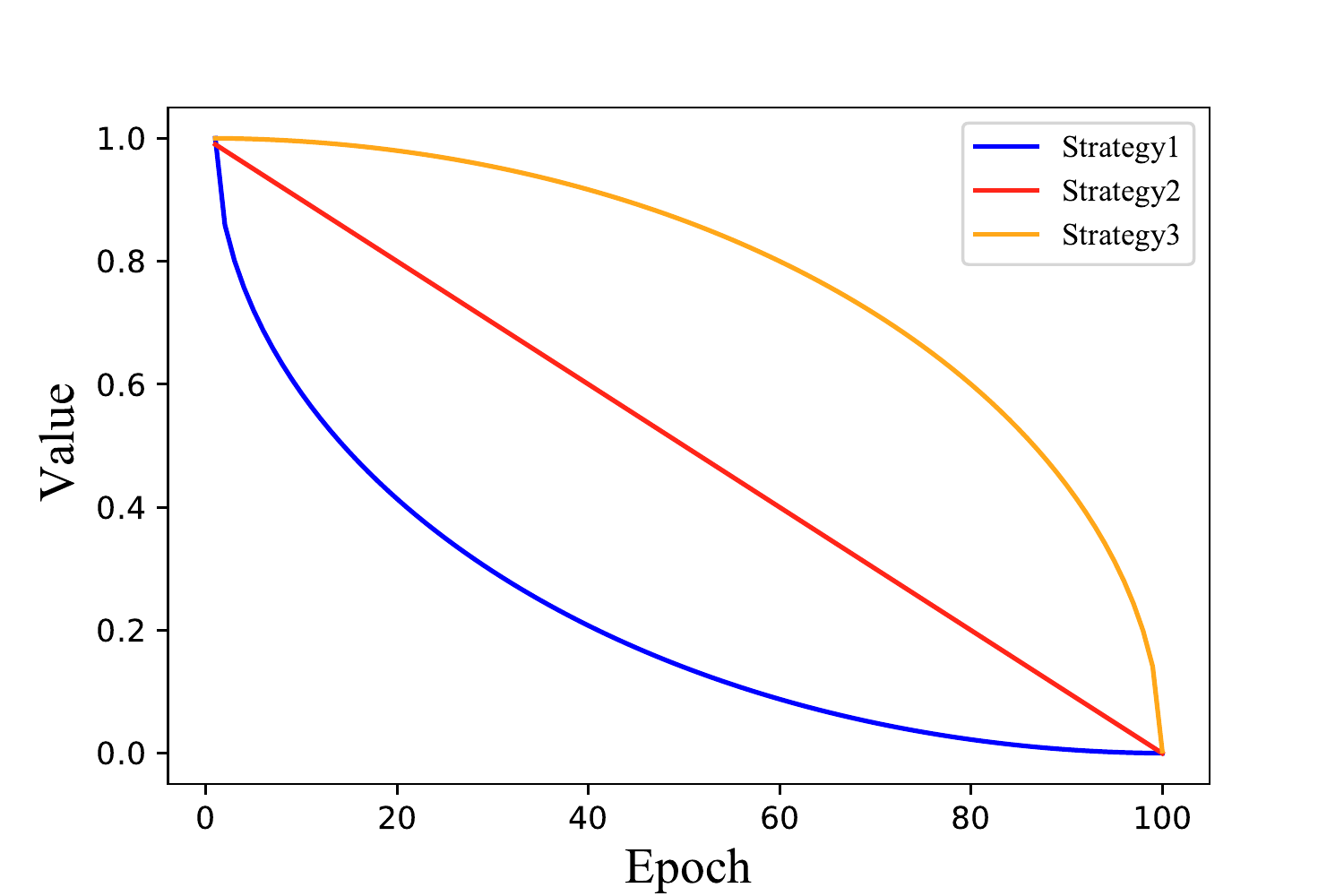}
	\caption{Three different strategies for $\lambda$, which controls the mixing ratio of second-order and first-order methods.}
	\label{fig:fig2}
\end{figure}

Then, we analyze the impact of different $\lambda$ parameter strategies on ANG performance, as shown in Fig. \ref{fig:fig2}. 
Correspondingly, through the resultsin Fig. \ref{fig:fig1}, we find that the more "concave" strategy 1 is closer to the first-order generalization performance on loss, and has the best suppression of overfitting. In terms of accuracy, strategy 1 also achieved the most stable and excellent performance in the both two data sets.

\subsubsection{Part Two} 
\begin{figure}[htbp]
	\centering
	\subfigure[{Validation Loss vs. Epoch}]{
		\includegraphics[scale=0.5]{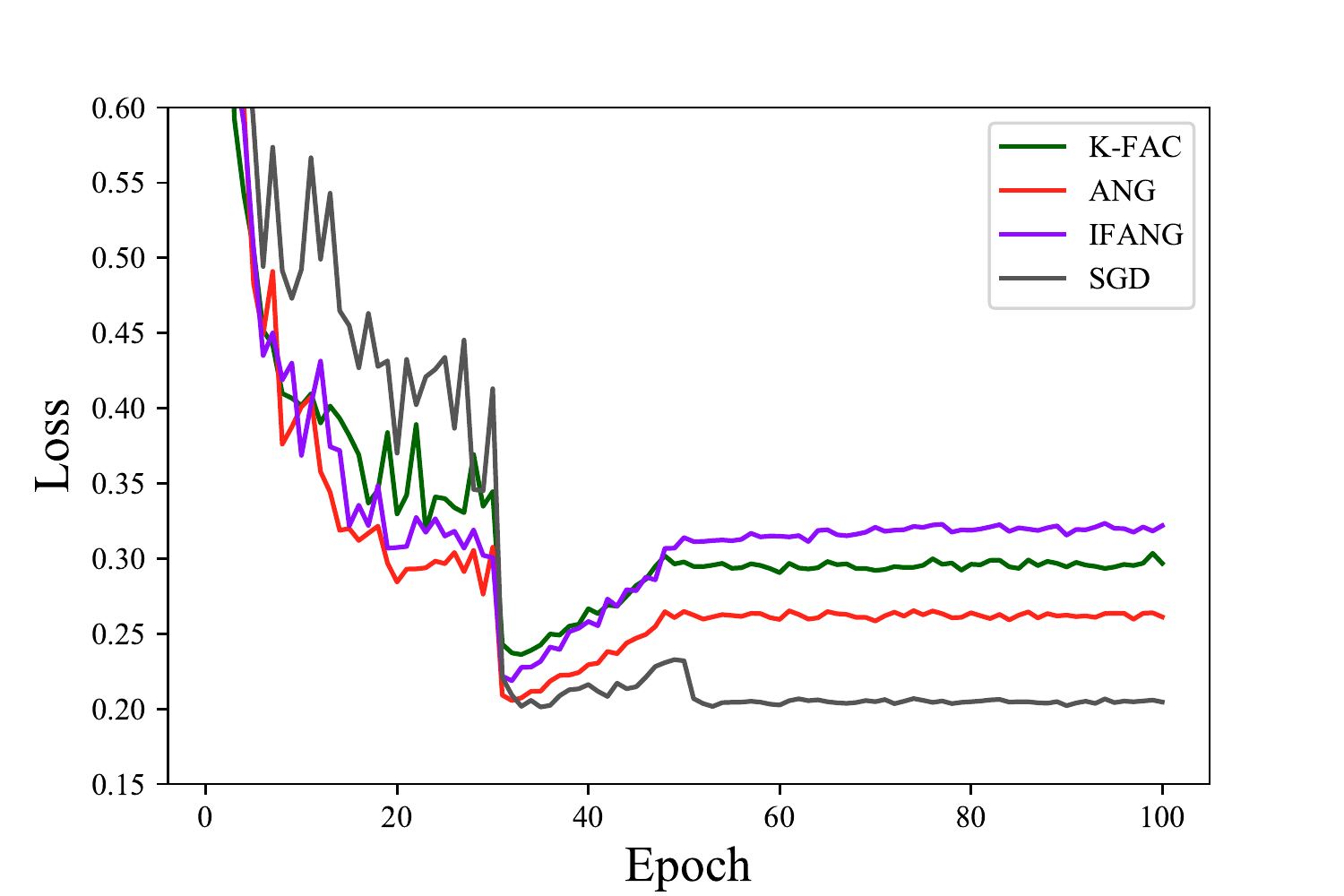}
	}
	\subfigure[{Validation Accuracy vs. Epoch}]{
		\includegraphics[scale=0.5]{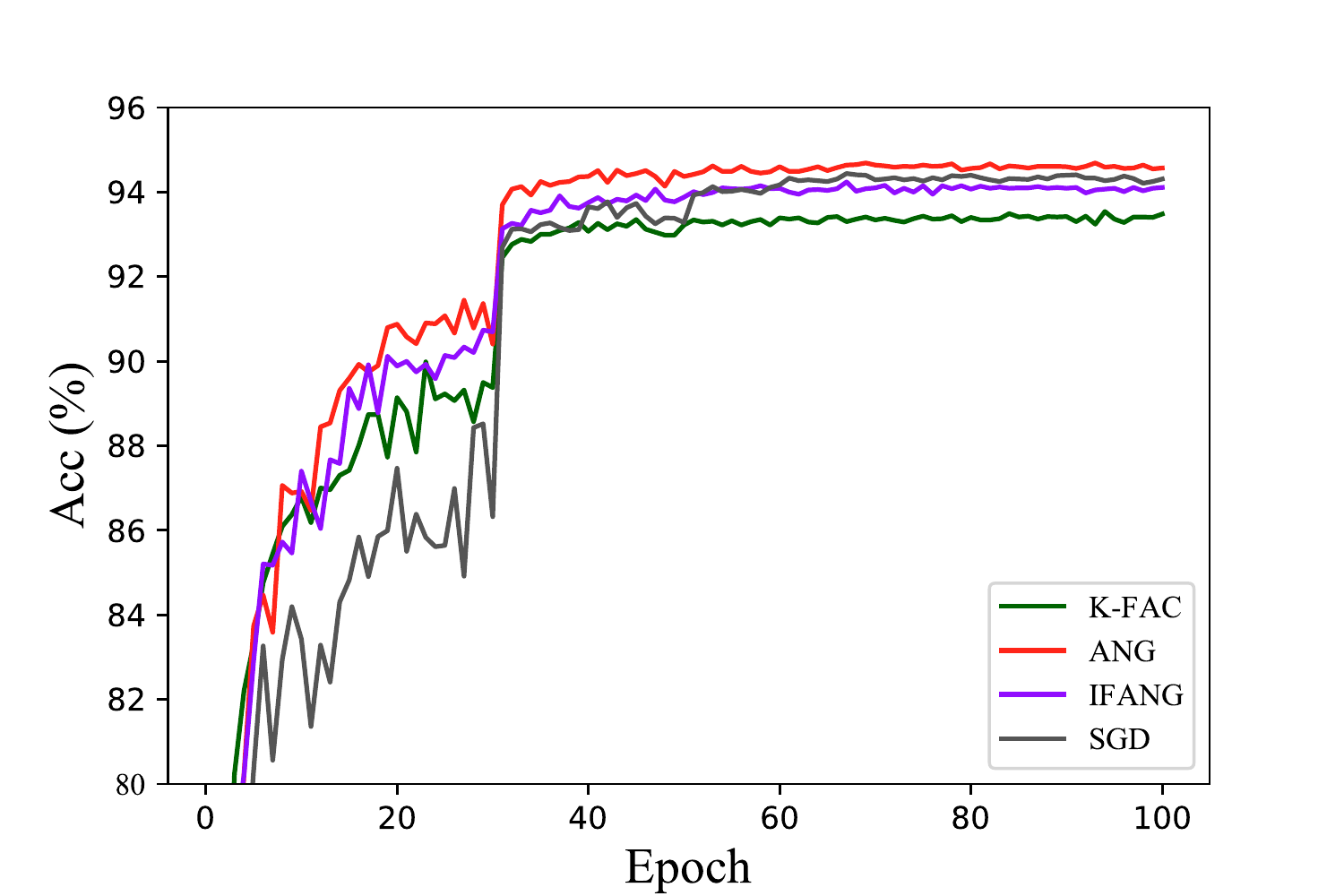}
	}
	\subfigure[{Validation Loss vs. Epoch}]{
		\includegraphics[scale=0.5]{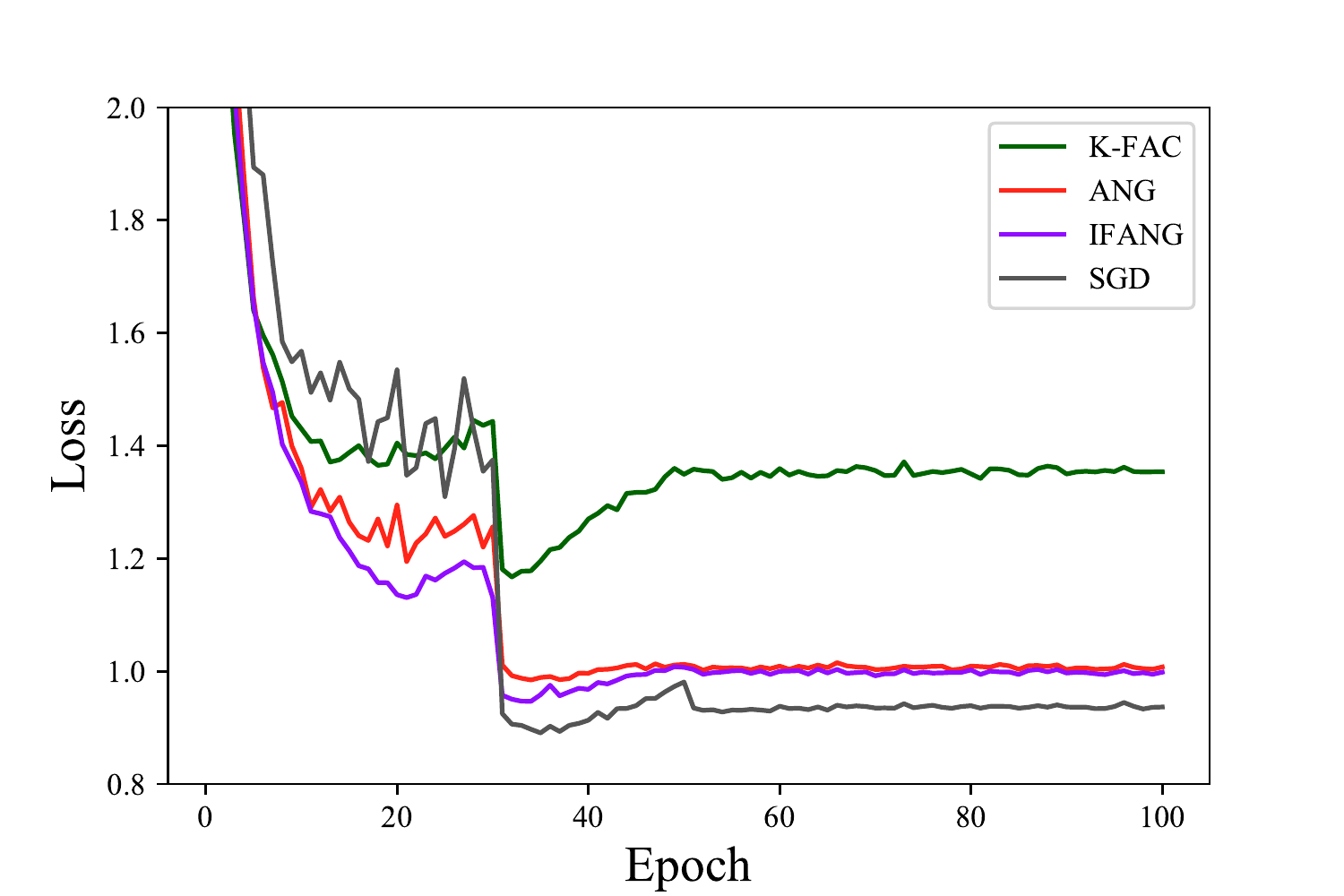}
	}
	\subfigure[{Validation Accuracy vs. Epoch}]{
		\includegraphics[scale=0.5]{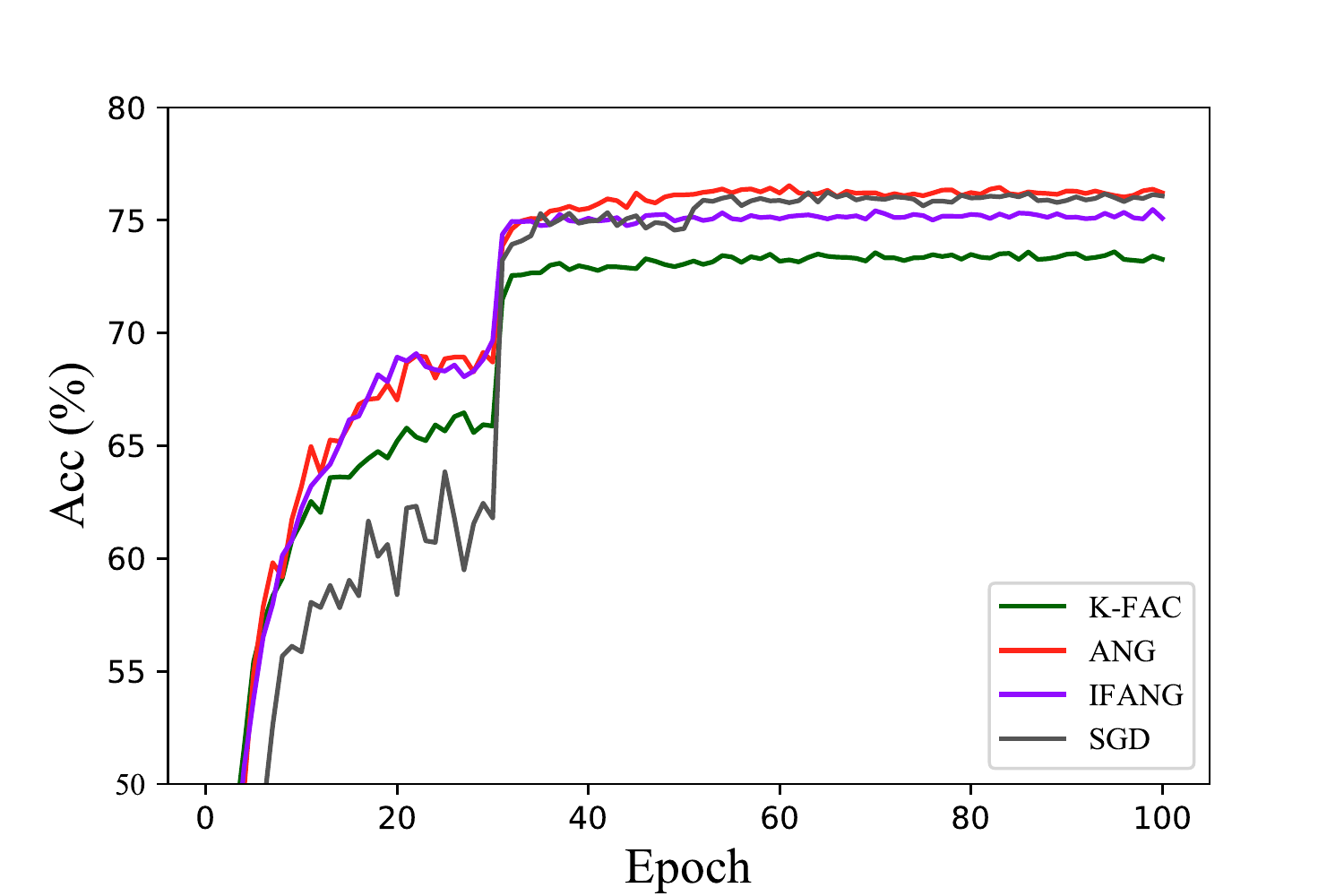}
	}
	\caption{Performance comparison among SGD, K-FAC, ANG and Truncated K-FAC on CIFAR-10 and CIFAR-100. \textbf{(a)}, \textbf{(b)} show the accuracy and loss curves on CIFAR-10 respectively, and \textbf{(c)}, \textbf{(d)} on CIFAR-100.}
	\label{fig:fig3}
\end{figure}

In this part, we focus on analyzing the performance of IFANG. In order to solve the inverse problem in second-order optimization, we introduce a 1-rank natural gradient estimation method based on ANG. ANG adopted more extreme sampling strategy. In view of the convolution and full connection layer, by compressing of feature dimension to 1, it gets 1 rank activation layer and layer gradient subspace, and by adopting the technology of SMW method, the dimension of factorized matrix can be reduced to 1. Finally, The calculation of the inverse matrix is reduced to the inverse of a scalar. Due to further estimation, some curvature information is lost.
IFANG delivers slightly less impressive but acceptable performance of accuracy and loss, as shown in Fig. \ref{fig:fig3}. However, we counted the wall-clock time of different algorithms and found IFANG's amazing advantage in speed. The detailed iteration times are exhibited in Table \ref{tab:tab1}. Specifically, from the perspective of the time of each epoch, IFANG reduced by 23\% compared with K-FAC. For a single iteration involving inversion, IFANG performed almost as well as SGD and was about 90\% more efficient than K-FAC. As for ANG, it adds some extra steps to integrate the gradient compared to k-FAC, thus slightly increasing the time.

\begin{table}[htpb]
	\caption{Time consumptions of different methods} 
	\centering
	\begin{tabular}{@{}ccccc@{}}
		\toprule
		Algorithm & Time consumption per iteration($s$) & Time Gap & Time consumption per epoch($s$) & Time Gap \\ \midrule
		K-FAC & 4.77   & 0\%      & 56.34 & 0\%      \\
		SGD   & 0.0144 & -99.70\% & 32.72 & -41.92\% \\
		ANG   & 4.855  & +1.78\%  & 56.48 & +0.24\%  \\
		IFANG & 0.341  & -92.85\% & 43.2  & -23.32\% \\ \bottomrule
	\end{tabular}
	\label{tab:tab1}
\end{table}

\section{Conclusion}
From the perspective of generalization performance, the proposed ANG provides a smooth and progressive regularization strategy, which enables the training to learn quickly in the early stage at a second-order speed and converge in the later stage at a first-order generalization performance, without any oscillation caused by mode collapse. In addition, to solve the problem of hard inverse in the second order, we further proposed a natural gradient algorithm IFANG without demand-inverse based on ANG. This algorithm is a more friendly alternative to different devices when the computing power is limited.

\bibliographystyle{unsrtnat}
\bibliography{references}  






\end{document}